\documentclass[conference]{IEEEtran}
\IEEEoverridecommandlockouts
\usepackage{cite}
\usepackage{amsmath,amssymb,amsfonts}
\usepackage{algorithmic}
\usepackage{graphicx}
\usepackage{textcomp}
\usepackage{xcolor}
\usepackage{amssymb} 
\usepackage{url}

\def\BibTeX{{\rm B\kern-.05em{\sc i\kern-.025em b}\kern-.08em
    T\kern-.1667em\lower.7ex\hbox{E}\kern-.125emX}}

\begin{document}

% \title{WeBe Band: On-Device Low-Latency Motion Recognition \\
% \title{MOTION: ML-Assisted On-device Triaxial Inference for On-the-fly Navigation with Low-Latency Motion Recognition\\
\title{MOTION: ML-Assisted On-Device Low-Latency Motion Recognition\\
% should not be used} (should we delete this line?)
}

\author{
    Veeramani Pugazhenthi$^{1}$, Wei-Hsiang Chu$^{2}$, Junwei Lu$^{2}$, Jadyn N. Miyahira$^{2}$, 
    Mahdi Eslamimehr$^{3}$\\
    Pratik Satam$^{4}$, Rozhin Yasaei$^{5}$, Soheil Salehi$^{1}$\\%
    $^{1}$Department of Electrical and Computer Engineering, University of Arizona, Tucson, AZ, U.S.A.\\ %
    $^{2}$Department of Computing and Data Science, University of California, Berkeley, Berkeley, CA, U.S.A.\\%
    $^{3}$Quandary Peak Research, Los Angeles, CA, U.S.A.\\
    $^{4}$Department of Systems and Industrial Engineering, University of Arizona, Tucson, AZ, U.S.A.\\
    $^{5}$College of Information Sciences, University of Arizona, Tucson, AZ, U.S.A.\\
    Email: \{$^{1}$veerpugazh5, $^{4}$pratiksatam, $^{5}$yasaei, $^{1}$ssalehi\}@arizona.edu, \\
    \{$^{2}$ruchu, $^{2}$jw.l, $^{2}$jadyn.miyahira\}@berkeley.edu,
    $^{3}$mahdi@quandarypeak.com%
}

\maketitle

\begin{abstract}

The use of tiny devices capable of low-latency gesture recognition is gaining momentum in everyday human-computer interaction and especially in medical monitoring fields. Embedded solutions such as fall detection, rehabilitation tracking, and patient supervision require fast and efficient tracking of movements while avoiding unwanted false alarms. 
This study presents an efficient solution on how to build very efficient motion-based models only using triaxial accelerometer sensors. We explore the capability of the AutoML pipelines to extract the most important features from the data segments. This approach also involves training multiple lightweight machine learning algorithms using the extracted features. We use \emph{WeBe Band}, a multi-sensor wearable device that is equipped with a powerful enough MCU to effectively perform gesture recognition entirely on the device. 
Of the models explored, we found that the neural network provided the best balance between accuracy, latency, and memory use. Our results also demonstrate that reliable real-time gesture recognition can be achieved in \emph{WeBe Band}, with great potential for real-time medical monitoring solutions that require a secure and fast response time.

\end{abstract}


\section{Introduction}

Motion recognition capability is one of the prominent applications that many wearable devices need to have for activity tracking. In medical settings, such as hospitals and clinics, accurate, continuous, and reliable monitoring is essential for applications ranging from patient rehabilitation to elderly fall detection \cite{fall-detection}, and device adherence. All such applications pose real challenges when deployed in sensitive environments \cite{b9}. Medical environments rapidly evolve, and conventional camera-based methods are prone to mistakes in recognizing complicated scenarios and unpredictable environments. Such systems are often too expensive, not secure, uncomfortable, infrastructure-dependent, and unsuitable for long-term or ambulatory use \cite{b10}. Most embedded motion detection algorithms rely on 6-channel gyroscope and accelerometer data (IMU). However, here we focus on accelerometer-based sensing to provide a more compact, non-invasive alternative that captures subtle motion dynamics at low cost.

\begin{figure}[t]
    \centering
    \includegraphics[width=0.8\columnwidth]{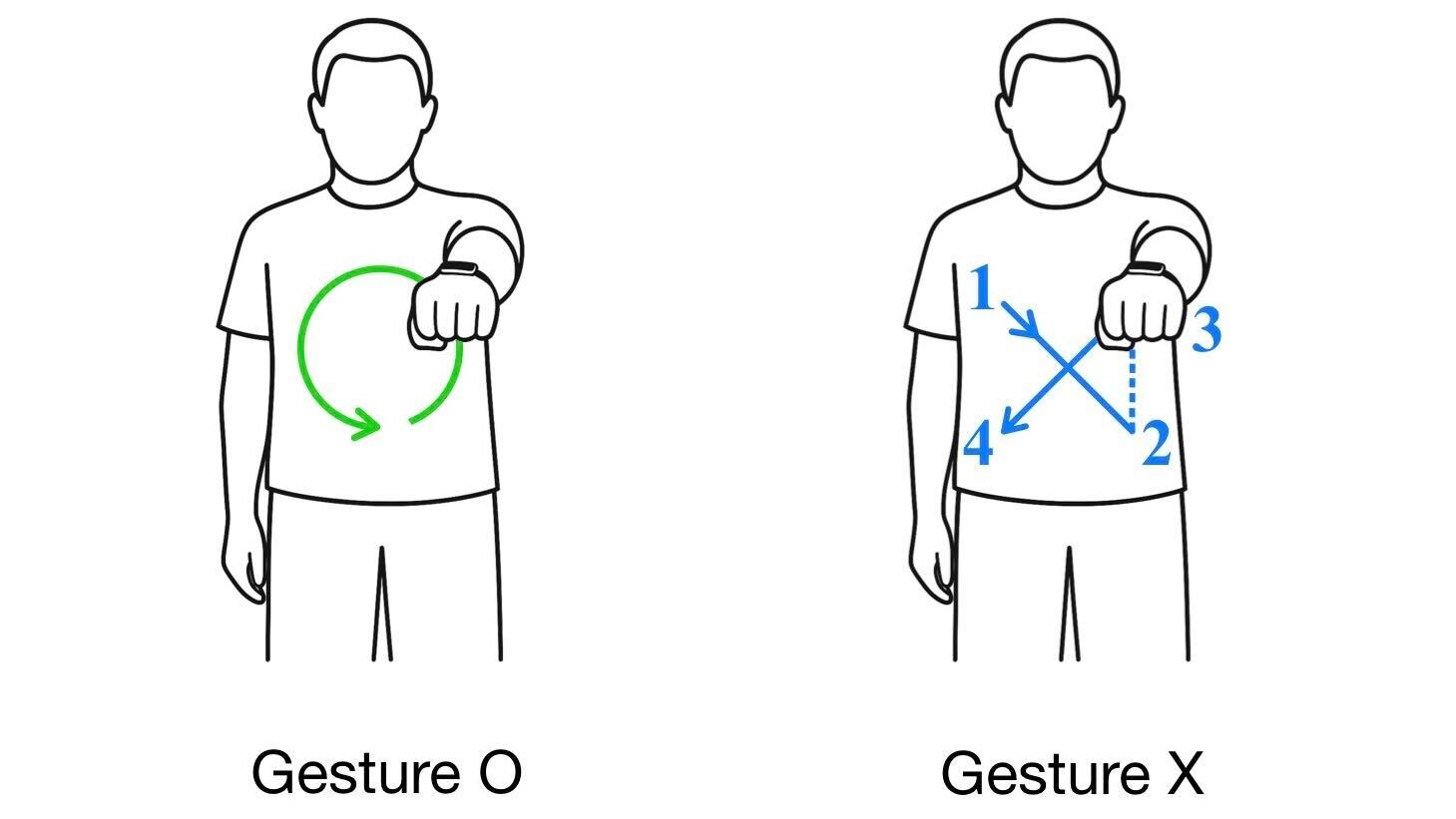}
    \caption{Illustration of the ``X'' and ``O'' gestures from the perspective of an external observer. The numbers indicate the sequential movement steps performed by the participant.}
    \label{fig:xo_gestures}
\end{figure}

Developing any embedded motion detection solution requires extra attention. Many approaches rely on handcrafted features extracted from 6-axis IMU signals and neglect the richer information that might be available in multi-axis accelerometer signals \cite{b12}. Limiting the signals to accelerometers does not necessarily indicate that there is not enough inherent information to operate a reliable motion recognition system. Improper data splitting or validation strategies can cause leakage, leading to overly optimistic performance estimates. Any embedded model needs to be tested using a simulator or on the device to validate its performance in the real setup.

We propose a generalizable gesture recognition framework using 3-axis accelerometer signals on \emph{WeBe Band}, a recently introduced watch-like wearable platform for continuous motion recognition \cite{webe_pubmed}. At its core, \emph{WeBe Band} is equipped with an ARM Cortex-M4F microcontroller ({\it nRF52840}, 64 MHz), with Flash and RAM sizes of 1 MB and 256 KB, respectively \cite{webe_pubmed,fang2024webe}.

In this study, we leverage the triaxial accelerometer signals of \emph{WeBe Band} to build a gesture detection algorithm, and use AutoML to extract the most informative features and augment the data to increase model robustness against unforeseen factors \cite{salehin2024automl}. We train and evaluate lightweight models, i.e. Neural Networks (NN), Random Forests (RF), Bonsai, and Pattern Matching Engine (PME), through device-level profiling. All developed gesture recognition models are deployed and executed directly on the device, with low-latency responses that are critical for applications such as fall detection and urgent medical alerts. On the other hand, raw sensor data never leaves the device; hence, enhancing privacy and security in super-sensitive medical environments.

%Our results show that high-performance gesture classification can be achieved with compact models that solely rely on accelerometers. The same approach, when combined with other sensors available on the \emph{WeBe Band}, i.e., photoplethysmography (PPG), electrodermal activity (EDA), and skin temperature, makes the \emph{WeBe Band} a promising platform for medical monitoring, rehabilitation tracking, elderly care, and real-time patient supervision in both hospital and home settings.

\section{Methodology}

\subsection{Data}

For this study, we exclusively use the integrated triaxial accelerometer sensor in \emph{WeBe Band}. Gesture data were collected at a sampling rate of 25 Hz.

%and used directly for training and evaluating the recognition models. 

\subsection{Data Collection}

Timeseries data were collected using the \emph{WeBe Band} worn on the left wrist. Five participants were recruited for the study. Each participant completed five recording sessions for each gesture. In each session, they performed approximately ten gestures separated by rest intervals of 3-5 seconds.

%to minimize overlap between motion instances and to facilitate manual data annotation.

We considered two significantly distinct gestures (illustrated in Fig.~\ref{fig:xo_gestures}):  

\begin{itemize}
    \item \textbf{“X” Gesture}: The participant moves the hand in a large ``X'' shape in the air, following the following sequence: starting from the center, 
    (1) to the top right,  
    (2) to the bottom left,  
    (3) to the top left,  
    (4) to the bottom right, and finally returning to the center.  

    \item \textbf{“O” Gesture}: The participant makes a clockwise circular motion forming the letter ``O''.  

    \item \textbf{Random Gestures}: The participants also performed unconfined hand movements, excluding deliberate shapes ``X'' and ``O'', to improve model robustness against non-target motions and to lower misclassification rates. 
\end{itemize}

\begin{figure}[t]
    \centering
    \includegraphics[width=0.8\columnwidth]{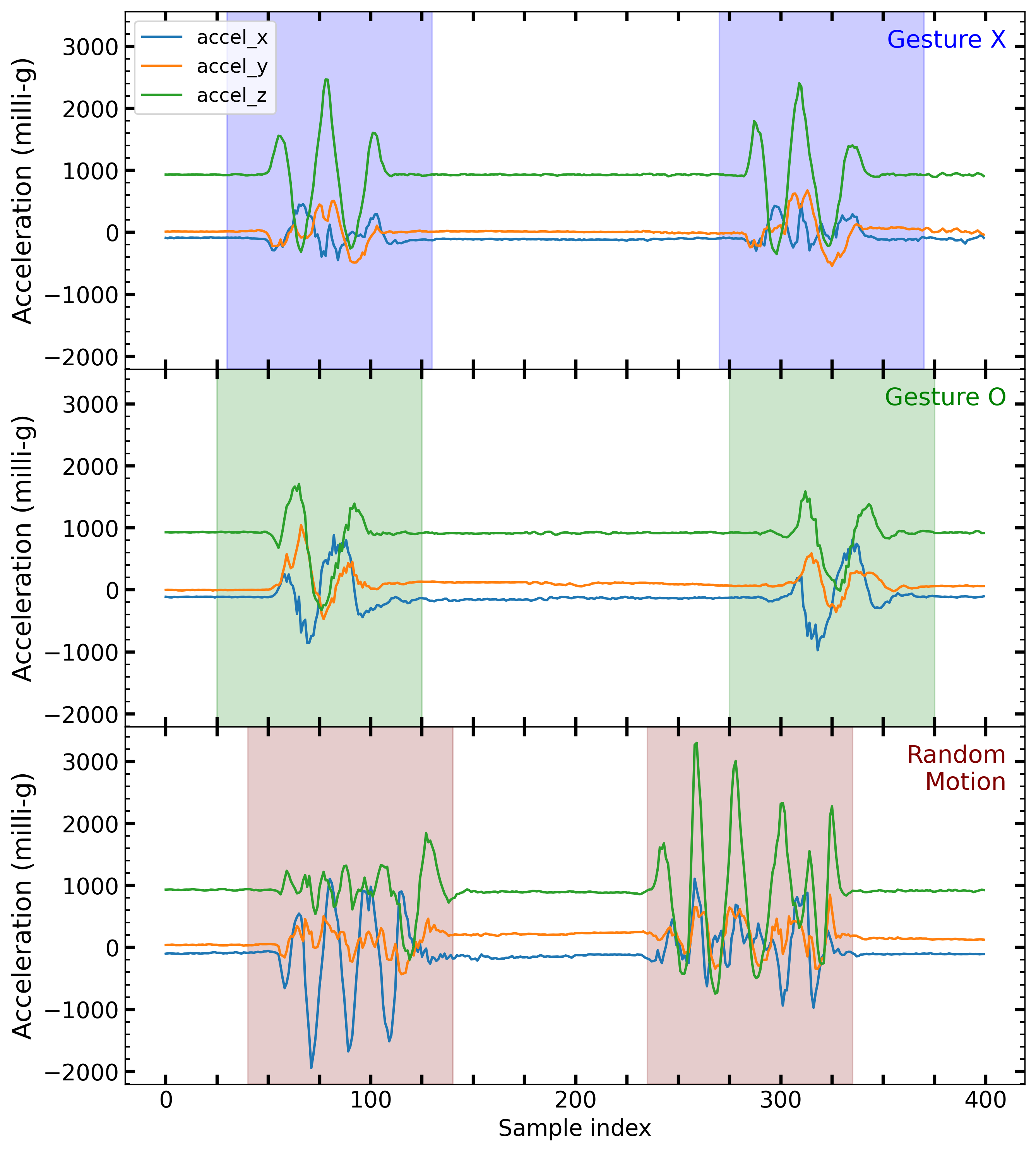}
    \caption{Representative tri-axial accelerometer signals for the three gesture classes, highlighted in color. From top to bottom (a) ``X,'' (b) ``O,'' and (c) Random Gestures. Each panel shows two instances of 4-sec gestures (highlighted in color), exhibiting the characteristic motion patterns that distinguish the classes.}
    \label{fig:gesture_signals}
\end{figure}

Representative accelerometer signals for ``X,'' ``O,'' and random gestures are shown in Fig.~\ref{fig:gesture_signals}.  

\subsection{Data Annotation}

We segmented all gestures, each containing $\gtrsim$100 samples ($\sim$4 seconds), and labeled them manually and semi-automatically. In some cases, we used the \emph{WeBe Band} event button to mark the exact moments of gestures, and used a custom script to extract and label segments. We manually verified the quality of all labels. 

\subsection{Data Augmentation}
\label{sec:data-augmentation}

To ensure balanced class representation between “X” and “O” gestures, we applied data augmentation crafted for timeseries data:

\begin{itemize}
    \item \textbf{Temporal Shifting:} In real-time deployment, data segments may be shifted depending on the state of the signal segmenter. Consequently, each label was randomly shifted by 7–15 samples to generate additional instances. 
    \item \textbf{Amplitude Scaling:} Signals were scaled by $\pm$10\% to simulate slight variations in gesture intensity and incorporate user-to-user variance in motion amplitude.
    \item \textbf{Time Stretch:} To address the variance of motion speed, signals were stretched by $\pm$5\%.
\end{itemize}

\section{ML Model Exploration}

\begin{figure}[t]
    \centering
    \includegraphics[width=.7\columnwidth]{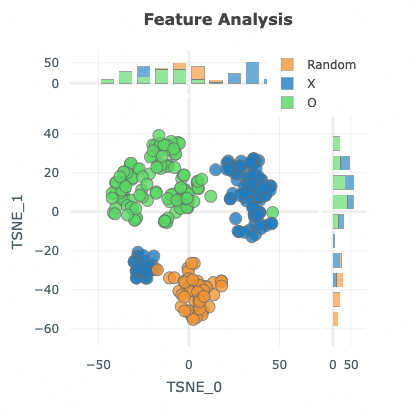}
    \caption{t-SNE visualization of the selected feature clusters for different gesture classes on the {\it WeBe Band} dataset.}
    \label{fig:tsne_features}
\end{figure}

\subsection{AutoML}

To identify promising features extracted from each segment, we first applied automated pipeline search using the open-source {\it Piccolo AI} toolkit \cite{piccolo}. It is designed to build and deploy ML models on embedded and IoT devices such as \emph{WeBe Band}. A key feature of this framework is its Automated Machine Learning (AutoML), that significantly helps researchers to systematically explore a large space of features, models, and hyperparameters. In a single run, the system evaluates a large number of candidates and produces a small set of models that perform well, along with the features that contributed to their success. {\it Piccolo AI} employs a genetic algorithm that evaluates an evolving set of extractors and classifiers based on accuracy, latency, and memory footprint. We mainly rely on this approach to find effective features, while it also offers insights good model candidates for embedded deployment. 

\subsection{Feature Selection}

Due to the random nature of AutoML, each run can return a different set of features/models. Hence, we repeated the process several times. For each iteration, we documented the feature sets used by best performing models, and documented the features used by these models. The feature pool we explored for the AutoML process included statistical descriptors (mean, variance, kurtosis), signal amplitude measures (peak-to-peak and min–max distance indicators, zero-crossing rates), and cross-axes feature fusion. The most common identified subset is summarized in Table~\ref{tab:features}, drawn from the feature sets of the top 5 well-performing models in all 4 iterations. To investigate the discriminative potential of these features, we used the t-SNE dimensionality reduction algorithm~\cite{vanDerMaaten2008}. Evidently, the selected features are capable of supporting reasonable classification (see Fig.~\ref{fig:tsne_features}).

\subsection{Model Generation}

We re-trained each individual algorithm on the best identified features. To maximize the models performances, we fine-tuned their hyperparameters.

\begin{table}[t]
\centering
\caption{Top feature candidates found by AutoML.}
\label{tab:features}
\resizebox{\columnwidth}{!}{
\begin{tabular}{l|c|c|c}
\hline
\textbf{Feature (with description)} & $accel-X$ & $accel-Y$ & $accel-Z$ \\
\hline \hline
Mean & \checkmark & \checkmark & - \\
Variance & - & \checkmark & \checkmark \\
Kurtosis & - & \checkmark & - \\
25th Percentile & - & \checkmark & - \\
75th-25th Percentile Range & \checkmark & \checkmark & - \\
Negative Zero Crossings & \checkmark & \checkmark & \checkmark \\
Global Min and Max Sum & \checkmark & - & - \\
Median Difference of Cross Axes & \checkmark & \checkmark & \checkmark \\
Min-Max Distances & \checkmark & \checkmark & \checkmark \\
Global Peak-to-peak Distance & - & \checkmark & \checkmark \\
\hline
\end{tabular}
}
\end{table}

We trained four ML algorithms as described below.

\begin{itemize}
    \item \textbf{Pattern Matching Engine (PME):} PME is a distance-based classifier that compares new inputs with the stored prototypes using a distance metric, e.g., $L_1$ or {\it L-sub} metrics \cite{hammer2011prototype}. Each prototype is associated with an \emph{Area of Influence (AIF)}, defining the maximum distance range within which a sample can be considered similar to that prototype. In our experiments, we set the maximum (minimum) allowed AIF to 400 (25), to balance specificity and robustness while avoiding overfitting. The maximum number of dynamically added neurons to span the entire feature space is set to 128.

    \item \textbf{Random Forest (RF):} Known as one of the strongest classical algorithms for motion recognition. Using multiple decision trees during training, it outputs the class by taking the majority vote among all predictions. This approach is usually well-suited for embedded applications and inherently exhibits robustness against overfitting, while being capable of effectively handling noisy sensor data. We trained a forest of 40 decision trees, each with a maximum depth of 7. These parameters were tuned to achieve the highest classification accuracy.

    \item \textbf{Bonsai:} The Bonsai tree optimizer is specifically designed for efficient inference on microcontrollers \cite{kumar2017bonsai}. It captures non-linear decision boundaries while maintaining a small footprint. Bonsai linearly maps high-dimensional feature spaces into lower dimensions in combination with a shallow decision tree. We optimized Bonsai with a set of fine-tuned hyperparameters, including a projection dimension of 13, tree depth of 4, and kernel width $\sigma=0.3$. The model was trained for 500 epochs with a batch size of 32 and a learning rate of 0.01.
    
\item \textbf{Neural Network (NN):} We trained several simple fully connected neural networks (NNs) on the extracted features. The network consists of four dense layers of varying sizes. After experimenting with a few architectures, we found the best-performing layer architecture of 16, 16, 8, and 4 neurons. Each NN was trained for 10 epochs with a batch size of 32 using the Adam optimizer, learning rate = 0.0015, dropout rate of 0.1 to prevent overfitting, and batch normalization. To reduce the false-positive rate, the output was post-processed with a conservative 60\% confidence threshold, mapping low-confidence predictions to ``Uncertain''.

\end{itemize}

\begin{table}[t]
\centering
\caption{Class-Specific Performance Metrics on test Set (\%)}
\label{tab:class_perf}
\resizebox{\columnwidth}{!}{%
\begin{tabular}{l|cc|cc|cc|cc}
\hline
\textbf{Gesture Class} & \multicolumn{2}{c|}{\textbf{PME*}} & \multicolumn{2}{c|}{\textbf{RF*}} & \multicolumn{2}{c|}{\textbf{Bonsai}} & \multicolumn{2}{c}{\textbf{NN*}} \\
\cline{2-9}
 & \textbf{Recall} & \textbf{Prec.} & \textbf{Recall} & \textbf{Prec.} & \textbf{Recall} & \textbf{Prec.} & \textbf{Recall} & \textbf{Prec.} \\
\hline \hline
O Gesture     & 99.1 & 98.2 & 95.2 & 95.2 & 95.2 & 97.5 & 99.1 & 96.5 \\
X Gesture     & 99.5 & 93.3 & 96.8 & 96.8 & 98.9 & 93.1 & 99.5 & 98.6 \\
Random Motion & 96.5 & 99.7 & 98.1 & 98.1 & 96.4 & 100 & 97.1 & 100 \\
\hline
Total Accuracy      & \multicolumn{2}{c|}{97.7} & \multicolumn{2}{c|}{97.6} & \multicolumn{2}{c|}{97.1} & \multicolumn{2}{c}{98.1} \\
Macro F1-score      & \multicolumn{2}{c|}{97.7} & \multicolumn{2}{c|}{96.8} & \multicolumn{2}{c|}{96.8} & \multicolumn{2}{c}{98.6} \\
\hline
\end{tabular}%
}
\\[4pt]
\footnotesize{*~PME = Pattern Matching Engine, RF = Random Forest, NN = Neural Network. Prec. = Precision, Recall = Recall (Precision)}
\end{table}

\section{Summary of Results}

We divided our data into train–validation–test sets with a fraction of 60-20-20\%. During the training and fine-tuning process, we class-balanced the data according to Section~\ref{sec:data-augmentation}. To realistically assess model misclassifications, we did not balance the test set to better reflect deployment conditions. Each model was assessed based on classification accuracy, F1-score, latency, and memory usage. We evaluated each model in class-specific {\it Precision} and {\it Recall}. Precision measures the percentage of generated predictions that are correct, while Recall measures the proportion of each ground-truth gesture that is recognized correctly. These metrics are summarized in Table~\ref{tab:class_perf}. Having a rigorous method to select the best features and hyperparameters tuning, their performance turned out to be insignificantly different, with the neural network performing the best overall. Table~\ref{tab:cm_nn_val} shows the confusion matrix of the neural network model in the test set. The model clearly recognizes target gestures and distinguishes them from random motions. Notably, only 3\% of the random ground truth segments were incorrectly mapped to other gestures. The overall performances of the investigated models are summarized in Table~\ref{tab:perf_summary}. Contrary to common assumptions, the neural network achieved both the highest accuracy and the lowest latency (1.2 ms). Although Random Forests, Bonsai, and PME were slower, they remain viable alternatives in low-memory scenarios.

\begin{table}[t]
\centering
\caption{Neural Network Confusion Matrix For the Test Set (rows = ground truth, columns = predictions).}
\label{tab:cm_nn_val}
\resizebox{\columnwidth}{!}{%
\begin{tabular}{c|c|c|c|c|c|c}
\hline
 & \textbf{O} & \textbf{X} & \textbf{Rand} & \textbf{UNC} & \textbf{Support} & \textbf{Recall (\%)} \\
\hline
\textbf{True O}    & 113 & 0   & 0   & 1   & 114 & 99.1 \\
\textbf{True X}    & 0   & 223 & 0   & 1   & 224 & 99.5 \\
\textbf{True Rand} & 4   & 3   & 482 & 7   & 496 & 97.1 \\
\hline \hline
\textbf{Predicted}   & 117 & 226 & 482 & 9   & 834 & -- \\

\textbf{Precession (\%)} & 96.5 & 98.6 & 100 & -- & -- & Acc = 98.1 \\
\hline
\end{tabular}}
\\[2pt]
\footnotesize{*~Rand = Random gestures, UNC = Uncertain}
\end{table}

\subsection{Model Profiling on the WeBe Device}\label{AA}
We profiled each model on the {\it WeBe Band} to evaluate their real-time performance. {\it Piccolo AI} provides an estimate of the memory usage of each model and also facilitates quantization-aware training of neural networks, as well as quantization using TensorFlow Lite for deployment on microcontrollers. We implemented a few profiling units in the {\it WeBe} firmware. To reduce profiling overhead, we leveraged the CoreSight debug block of the Data Watchpoint and Trace (DWT) unit. A 32-bit cycle counter was sampled immediately before and after each model call. Using the clock frequency of 64 MHz, we converted the cycle difference into microseconds to precisely measure latency. Table~\ref{tab:perf_summary} presents the results. Surprisingly, the neural network achieves the lowest latency while still delivering the best overall performance, which highlights the benefits of our feature selection procedure. The PME model also appears promising for memory constraint systems.

\begin{table}[t]
\centering
\caption{Performance comparison of models and AutoML outcomes on the WeBe Band.}
\label{tab:perf_summary}
\renewcommand{\arraystretch}{1.2}
\normalsize
\resizebox{\columnwidth}{!}{%
\begin{tabular}{l|c|c|c|c|c}
\hline
\textbf{ML Model} & \textbf{Accuracy} & \textbf{F1-score} & \textbf{Latency} & \textbf{SRAM+STACK} & \textbf{FLASH} \\
\hline
 & (\%) & (\%) & (ms) & (KB) & (KB) \\
\hline \hline
AutoML$^*$ (NN$^{**}$) & 97.5 & 97.7 & 1.2 & 7.7 & 18.0\\
Random Forest           & 97.6 & 96.8 & 4.6 & 2.6 & 20.1\\
Bonsai                  & 97.1 & 96.8 & 5.7 & 3.1 & 17.5\\
PME                     & 97.7 & 97.7 & 4.6 & 2.7 & 9.8\\
\textbf{Neural Network (NN)}          & \textbf{98.1} & \textbf{98.6} & \textbf{1.2} & \textbf{7.7} & \textbf{18.0}\\
\hline
\end{tabular}}
\\[4pt]
\footnotesize{$^*$ AutoML = best pipeline identified. $^{**}$ NN = Neural Network. }
\end{table}

\section{Conclusion and Future Work}

% In this study, we demonstrated the feasibility of performing accurate low-latency gesture recognition directly on \emph{WeBe Band} utilizing only triaxial accelerometer sensors. We built compact machine learning models with carefully optimized selected features. All of our fine-tuned models achieved robust on-device classification and performed satisfactorily when deployed on the device. We found that among the models explored, neural networks not only achieved the lowest latency ($<2$ milliseconds), outperforming traditional machine learning approaches such as ensemble decision trees and distance-based classifiers. All models exhibited a better overall accuracy than 95\%, with neural networks showing promise to achieve higher performance, highlighting that with the right features and efficient hyper-parameter tuning, all of the explored models can remain compact and fast on microcontrollers. Beyond gesture recognition, the {\it WeBe Band} offers a versatile platform for medical and everyday health monitoring, with potential applications such as rehabilitation tracking, fall detection, and continuous patient supervision. Future work involves integrating additional biomarkers, as well as developing model triggering mechanisms to enable longer battery life when needed.

In this study, we demonstrated the feasibility of performing accurate low-latency gesture recognition directly on \emph{WeBe Band} utilizing only triaxial accelerometer sensors. We built compact machine learning models with carefully optimized selected features. All of our fine-tuned models achieved robust on-device classification and performed satisfactorily when deployed on the device. We found that among the models explored, neural networks not only achieved the lowest latency ($<2$ milliseconds), outperforming traditional machine learning approaches such as ensemble decision trees and distance-based classifiers. All models exhibited an overall accuracy better than 95\%, with neural networks showing promise to achieve higher performance. This result highlights that with the right features and efficient hyperparameter tuning, all of the explored models can remain both compact and fast on microcontrollers. 

%The importance of automated feature exploration and model hyper-tuning becomes more evident, as AutoML reveals its critical role in identifying discriminative features and providing insights into various promising models, paving the path toward further model refinements.

% ; however, the difference is not significant.

Beyond gesture recognition, the {\it WeBe Band} offers a versatile platform for medical and everyday health monitoring, with potential applications such as rehabilitation tracking, fall detection, and continuous patient supervision. Future work involves integrating additional biomarkers, such as PPG, as well as developing model triggering mechanisms to enable longer battery life and always-on operation when needed.

\section*{Acknowledgment}
We gratefully acknowledge the \emph{WeBe Band} team for their support in deploying our solutions on the device.

\end{document}